\documentclass[10pt,a4paper,twocolumn]{article}
\usepackage{f1000_styles}
\usepackage{amsmath}
\usepackage{latexsym}
\usepackage{listings}
\usepackage{url}

\newcommand{\simiid}{\stackrel{\mathrm{iid}}{\sim}}
\newcommand{\m}[1]{\mathrm{#1} }
\renewcommand{\cal}[1]{\mathcal{#1}}
\renewcommand{\v}[1]{\boldsymbol{#1}}
\newcommand{\bb}[1]{\mathbb{#1}}

\usepackage[numbers]{natbib}
\usepackage[ruled, noend]{algorithm2e}

\begin{document}

\title{Graph Neural Network-Based Anomaly Detection for River Network Systems}
\titlenote{}
\author[1,2]{Katie Buchhorn}
\author[1,2]{Edgar Santos-Fernandez}
\author[1,2]{Kerrie Mengersen}
\author[1,3]{Robert Salomone}

\affil[1]{Queensland University of Technology, School of Mathematical Sciences}
\affil[2]{Queensland University of Technology, Centre for Data Science}
\affil[3]{Queensland University of Technology, School of Computer Science}

\maketitle
\thispagestyle{fancy}

\begin{abstract}

Water is the lifeblood of river networks, and its quality plays a crucial role in sustaining both aquatic ecosystems and human societies.
Real-time monitoring of water quality is increasingly reliant on in-situ sensor technology.
Anomaly detection is crucial for identifying erroneous patterns in sensor data, but can be a challenging task due to the complexity and variability of the data, even under typical conditions.
This paper presents a solution to the challenging task of anomaly detection for river network sensor data, which is essential for accurate and continuous monitoring. 
We use a graph neural network model, the recently proposed Graph Deviation Network (GDN), which employs graph attention-based forecasting to capture the complex spatio-temporal relationships between sensors. 
We propose an alternate anomaly threshold criteria for the model, GDN+, based on the learned graph.
To evaluate the model's efficacy, we introduce new benchmarking simulation experiments with highly-sophisticated dependency structures and subsequence anomalies of various types.
We further examine the strengths and weaknesses of this baseline approach, GDN, in comparison to other benchmarking methods on complex real-world river network data.
Findings suggest that GDN+ outperforms the baseline approach in high-dimensional data, while also providing improved interpretability.
We also introduce software called {\fontfamily{qcr}\selectfont gnnad}.

\end{abstract}

\section*{Keywords}
Anomaly Detection, Graph Deviation Network, Graph Neural Network, Multivariate Time Series, Graph Attention Forecasting, Spatio-temporal Data, Complex Systems

\clearpage

\section*{Introduction}

River network systems play a vital role as freshwater habitats for aquatic life, and as support for terrestrial ecosystems in riparian zones, but are particularly sensitive to the anthropogenic impacts of climate change, water pollution and over-exploitation, among other factors.
As a United Nations Sustainable Development Goal \cite{assembly2015transforming}, water quality is a major environmental concern worldwide.
The use of \textit{in-situ}\footnote{in-situ refers to an instrument in direct contact with the medium of observation.} sensors for data collection on river networks is increasingly prevalent \cite{silva2022advances, ritchie2003remote}, generating large amounts of data that allow for the identification of fine-scale spatial and temporal patterns, trends, and extremes, as well as potential sources of pollutants and their downstream impacts.
However, such sensors are susceptible to technical errors relating to the equipment, herein defined as \textit{anomalies}, for example due to miscalibration, biofouling, electrical interference and battery failure.
In contrast, extreme {\em events} in rivers occur as result of heavy rain and floods.
Technical anomalies must be identified before the data are considered for further analysis, as they can introduce bias in model parameters and affect the validity of statistical inferences, confounding the identification of true changes in water variables.
Trustworthy data is needed to produce reliable and accurate assessments of water quality, for enhanced environmental monitoring, and for guiding management decisions in the prioritisation of ecosystem health.

Anomaly detection in river networks is challenging due to the highly dynamic nature of river water even under typical conditions \cite{kang2009discovering}, as well as the complex spatial relationships between sensors.
The unique spatial relationships between neighbouring sensors on a river network are characterised by a branching network topology with flow direction and connectivity, embedded within the 3-D terrestrial landscape.
Common anomalies from data obtained from in-situ sensors are generally characterised by multiple consecutive observations ({\em subsequence} or persistent, \cite{blazquez2021review}), including sensor drift and periods of unusually high or low variability, which may indicate the necessity for sensor maintenance or calibration \cite{bourgeois2003use}.
Such anomalies are difficult to detect and often associated with high false negative rates \cite{leigh2019framework}.

Earlier statistical studies have focused on developing autocovariance models based on within-river relationships to capture the unique spatial characteristics of rivers \cite{ver2010moving}.
Although these methods are adaptable to different climate zones, and have recently been extended to take temporal dependencies into account \cite{santos2022bayesian}, data sets generated by in-situ sensors still pose significant computational challenges with such prediction methods due to the sheer volume of data \cite{porter2012staying}. Autocovariance matrices must be inverted when fitting spatio-temporal models and making predictions, and the distances between sites must be known.
Previous work \cite{rodriguez2020detecting}, aimed to detect drift and high-variability anomalies in water quality variables, by studying a range of neural networks calibrated using a Bayesian multi-objective optimisation procedure.
However, the study was limited to analyzing univariate time series data, and the supervised methods required a significant amount of labeled data for training, which are not always available.

There are limited unsupervised anomaly detection methods for subsequence anomalies (of variable length), and even less so for multivariate time series anomaly detection \cite{blazquez2021review}.
One such method uses dynamic clustering on learned segmented windows to identify global and local anomalies \cite{wang2018exact}.
However, this algorithm requires the time series to be well-aligned, and is not suitable for the lagged temporal relationships observed with river flow.
Another method to detect variable-length subsequence anomalies in multivariate time series data uses dimensionality reduction to construct a one-dimensional feature, to represent the density of a local region in the recurrence representation, indicating the recurrence of patterns obtained by a sliding window \cite{hu2019novel}.
A similarity measure is used to classify subsequences as either non-anomalous or anomalous.
Only two summary statistics were used in this similarity measure and the results were limited to a low-dimensional simulation study.
In contrast, the technique introduced by \cite{munir2018deepant}, DeepAnT, uses a deep convolutional neural network (CNN) to predict one step ahead. 
This approach uses Euclidean distance of the forecast errors as the anomaly score.
However, an anomaly threshold must be provided.

Despite the above initial advances, challenges still remain in detecting persistent variable-length anomalies within high-dimensional data exhibiting noisy and complex spatial and temporal dependencies.
With the aim of addressing such challenges, we explore the application of the recently-proposed Graph Deviation Network (GDN) \cite{deng2021graph}, and explore refinements with respect to anomaly scoring that address the needs of environmental monitoring.
The GDN approach \cite{deng2021graph} is a state-of-the-art model that uses sensor embeddings to capture inter-sensor relationships as a learned graph, and employs graph attention-based forecasting to predict future sensor behaviour.
Anomalies are flagged when the error scores are above a calculated threshold value. 
By learning the interdependencies among variables and predicting based on the typical patterns of the system in a semi-supervised manner, this approach is able to detect deviations when the expected spatial dependencies are disrupted.
As such, GDN offers the ability to detect even the small-deviation anomalies generally overlooked by other distance based and density based anomaly detection methods for time series \cite{schmidl2022anomaly, wilkinson2017visualizing}, while offering robustness to lagged variable relationships.
Unlike the commonly-used statistical methods that explicitly model covariance as a function of distance, GDN is flexible in capturing complex variable relationships independent of distance.
GDN is also a semi-supervised approach, eliminating the need to label large amounts of data, and offers a computationally efficient solution to handle the ever increasing supply of sensor data.
Despite the existing suite of methods developed for anomaly detection, only a limited number of corresponding software packages are available to practitioners.
In summary, we have identified the following gaps in the current literature and research on this topic: 

\begin{enumerate}
    \item An urgent need exists for a flexible approach that can effectively capture complex spatial relationships in river networks without the specification of an autocovariance model, and the ability to learn from limited labeled data, in a computationally efficient manner.
    \item Lack of data and anomaly generation schemes on which to benchmark methods, that exhibit complex spatial and temporal dependencies, as observed across river networks.
    \item Lack of open-source software for anomaly detection, which hinders the accessibility and reproducibility of research in this field, and limits the ability for individuals and organisations to implement effective anomaly detection strategies.
\end{enumerate}

Our work makes four primary contributions to the field:
\begin{enumerate}
    \item An improvement of the GDN approach via the threshold calculation based on the learned graph is presented, and shown to detect anomalies more accurately than GDN while improving the ability to locate anomalies across a network. 
    \item Methods for simulating new benchmark data with highly-sophisticated spatio-temporal structures are provided, contaminated with various types of persistent anomalies.
    \item Numerical studies are conducted, featuring a suite of benchmarking data sets, as well as real-world river network data, to explore the strengths and limitations of GDN (and its variants) in increasingly challenging settings. 
    \item User-friendly, free open-source software for the GDN/GDN+ approach is made available on the pip repository as {\fontfamily{qcr}\selectfont gnnad}, with data and anomaly generation modules, as well as the publication of a novel real-world data set.
\end{enumerate}

The structure of the remainder of the paper is as follows: The next section details the methods of GDN and the model extension GDN+, and describes the methodology of the simulated data and anomaly generation.
In the Results section we present an extensive simulation study on the benchmarking data, as well as a real-world case study.
The performance of GDN/GDN+ is assessed against other state-of-the-art anomaly detection models.
Further details and example code for the newly-released software are also provided.
The paper concludes with a discussion of the findings, and the strengths and weaknesses of the considered models.

\section*{Methods}

Consider multivariate time series data $\m Y = \left[\v y^{(1)}, \ldots, \v y^{(T)}\right]$, obtained from $n$ sensors over $T$ time ticks.
The (univariate) data collected from sensor $i=1,\ldots,n$ at time $t=1,\ldots,T$ are denoted as $y_i^{(t)}$.
Following the standard semi-supervised anomaly detection approach \cite{goldstein2016comparative, nassif2021machine}, non-anomalous data are used for training, while the test set may contain anomalous data.
That is, we aim to learn the sensor behaviour using data obtained under standard operational conditions throughout the training phase and identify anomalous sensor readings during testing, as those which deviate substantially from the learned behaviour.
As the algorithm output, each test point $\v y^{(t)}$ is assigned a binary label, $a(t) \in \{0, 1\}$, where $a(t) = 1$ indicates an anomaly at time $t$, anywhere across the full sensor network.

The GDN approach \cite{deng2021graph} for anomaly detection is composed of two aspects: 
\begin{enumerate}
    \item \textbf{Forecasting-based time series model:} a non-linear {\em autoregressive} multivariate time series model that involves graph neural networks is trained, and  
    \item \textbf{Threshold-based anomaly detection:} transformations of the individual forecasting errors are used to determine if an anomaly has occurred, if such errors exceed a calculated threshold. 
\end{enumerate}
The above components are described in more detail below. \\

\textbf{Forecasting-based Time Series Model.} To predict $\v y^{(t)}$, the model takes as input $w \in \bb N$ lags of the multivariate series,
\[\m X^{(t)} := \left[\v y^{(t-1)}, \ldots, \v y^{(t-w)}\right].\]
The $i$-th row (containing sensor $i$'s measurements for the previous $w$ lags) of the above input matrix is represented by the column vector, $\v x_i^{(t)} = (y_{i}^{(t-1)}, \ldots, y_{i}^{(t-w)})$.
Prior to training, the practitioner specifies acceptable {\em candidate relationships} via the sets $\mathcal{C}_1,\ldots, \mathcal{C}_n$, where each $\mathcal{C}_i \subseteq \{1,2,\ldots, n\}$ and does not contain $i$.
These sets specify which nodes are allowed to be considered to be connected {\em from} node $i$ (noting that the adjacency graph connections are not necessarily symmetric). 

The model implicitly learns a graph structure via training {\em sensor embedding} parameters $\v v_i \in \bb R^{d}$ for $i = 1,\ldots, n$ which are used to construct a graph. The intuition is that the embedding vectors capture the inherent characteristics of each sensor, and that sensors which are ``similar'' in terms of the angle between their vector embeddings are considered connected. 
Formally, the quantity $e_{ji}$ is defined as the cosine similarity between the vector embeddings of sensors $i$ and $j$:
\[e_{ji} = \frac{\v v_i^\top \v v_j}{||\v v_i||\, ||\v v_j||} \bb I\{j \in {\cal C}_i \}, \quad i,j \in \{1, \ldots, n \}, \] 
with $||\cdot ||$ denoting the Euclidean norm, and indicator function $\bb I$ which equals 1 when node $j$ belongs to set ${\cal C}_i$, and 0 otherwise.
Note that the similarity is forced to be zero if a connecting node is not in the permissible candidate set.
Next, let $e_{j,(i)}$ be the $i$-th largest value in $(e_{j1}, \ldots, e_{jn})$.
A {\em graph-adjacency matrix} (and in turn a graph itself) is then constructed from the sensor similarities via:
$$A_{ji} = \bb I\{\{e_{ji} \ge e_{{j,(K)}} \} \cup \{i=j \}\},$$ 
for user-specified $K \in \{1,\ldots, n\}$ which determines the maximum number of edges from a node, referred to as the ``Top-K'' hyperparameter.

The above describes how the trainable parameters $\{ \v v_k\}_{k=1}^n$ yield a graph. 
Next, the lagged series are fed individually through a shallow {\em Graph Attention Network} \citep{velivckovic2017graph} that uses the previously constructed graph.
Here, each {\em node} corresponds to a sensor, and the {\em node features} for node $i$ are the lagged (univariate) time-series values, $\v x_i^{(t)} \in \bb R^{w}$. Allow a parameter weight matrix $\m W \in \bb R^{d \times w}$ to apply a shared linear transform to each node. Then, the output of the network is given by
\[\v z_i^{(t)} = \max \left\{ \v 0, \left(\sum_{j: A_{ji}>0} \alpha_{ij} \m W \v x_i^{(t)}\right)\right\},\]
where $\v z_i^{(t)}$ is called the {\em node representation}, and coefficients $\alpha_{ij}$ are the attention paid to node $j$ when computing the representation for node $i$, with:

\begin{align}
    \pi_{ij} &= {\rm LeakyReLU}\left(\v a^\top \left(\v v_i \oplus \m W\v x_i^{(t)} + \v v_j \oplus \m W \v x_j^{(t)} \right)\right), \label{eq:pi} \\
    &\text{where }\alpha_{ij} = \frac{\exp(\pi_{ij})}{\sum_{k: A_{ki}>0}\exp(\pi_{ik})},\nonumber
\end{align}

with learnable parameters $\v a \in \bb R^{2d}$, where $\oplus$ denotes concatenation, and ${\rm LeakyReLU}(\v x) := \max\{\delta \v x, \v x\}$ for $\delta > 0$, with the maximum operation applied elementwise.
Note the addition\footnote{It seems that the authors of \cite{deng2021graph} mistakenly denote this addition as concatenation in the paper, however, their corresponding reference code computes addition.} in Equation \ref{eq:pi} and that $\sum_{j=1}^n\alpha_{ij}= 1$.
Intuitively, the above is an automated mechanism to aggregate information from a node itself and neighbouring nodes (whilst simultaneously assigning a weight of how much information to take from each neighbour) to compute a vector representing extracted information about node $i$ itself and its neighbours' interaction with it.
The final model output (prediction) is given by, 
\[\widehat{\v y}^{(t)} = f_{\v \eta}\left(\left[\v v_1 \odot \v z_1^{(t)}, \ldots, \v v_{n} \odot \v z_{n}^{(t)} \right] \right),\]
where $f_{\v \eta}: \bb R^{d \times n} \to \bb R^n$ is a feedforward neural network with parameters $\v \eta$, and $\odot$ denotes element-wise multiplication.
The model is trained by optimizing the parameters $\{ \v v_i\}_{i=1}^n$, $\m W$, $\v a$, and $\v \eta$ to minimize the mean squared error loss function
\[{\cal L} = \frac{1}{T-w}\sum_{t=w+1}^{T}\bigg|\bigg|\widehat{\v y}^{(t)} - \v y^{(t)} \bigg|\bigg|^2.\]
 
{\bf Threshold-based Anomaly Detection.} 
Given the learned inter-sensor and temporal relationships, we are able to detect anomalies as those which deviate from these interdependencies. 
An {\em anomalousness score} is computed for each time point in the test data. 
For each sensor $i$, we denote the prediction error at time $t$ as,
\[\epsilon_{i, t} = |y_i^{(t)} - \widehat{ y}_i^{(t)}|,\]
with $|\cdot|$ denoting the absolute value, and the vector of prediction error for each sensor is, $\v \epsilon_i \in \bb R ^{T-w}$. Since the error values of different sensors may vary substantially, we perform a robust normalisation of each sensor's errors to prevent any one sensor from overly dominating the others, that is,
\[\tilde{\v \epsilon}_i = \left(\frac{\v \epsilon_i - {\rm Median}(\v \epsilon_{i})}{{\rm IQR}(\v \epsilon_{i})}\right),\]
where IQR denotes {\em inter-quartile range}. In the original work by \cite{deng2021graph}, a time point $t$ is flagged as anomalous if,
\[A(t) = \bb I\{ \max_i (\tilde{\epsilon}_{i,t}) > \kappa\},\] 
using the notation $\tilde{\epsilon}_{i,t}$ for the error value at the $t$-th index.
Alternatively, the authors recommend using a {\em simple moving average} of $\tilde{\v \epsilon}_i$ and flagging time $t$ as anomalous if the maximum of that moving average exceeds $\kappa$.
The authors specify $\kappa$ as the maximum of the normalised errors observed on some (non-anomalous) validation data, denoted by variant epsilon, $\tilde{\v \varepsilon}$.
However, this is applied to all sensors as a fixed threshold.

\subsection*{Sensor-based Anomaly Threshold: GDN+}
The behavior of water quality variables may differ across space, for example, water level at high-altitude river network branches are generally characterised by rainfall patterns, whereas water level downstream near a river outlet can also be influenced by tidal patterns.
A fixed threshold across the network does not allow for any local adaptations in error sensitivity.
For this reason, this work also considers the novel sensor-specific threshold calculation,
\[A_i(t) = \bb I\{\tilde{\epsilon}_{i,t} > \kappa_i \},\]
where $\kappa_i$ is chosen such that,
\[\frac{1}{|\{\tilde{\varepsilon}_{j,t}\}_{j: A_{ji}>0}|}\sum_{j: A_{ji}>0} \bb I\{ \tilde{\varepsilon}_{j,t} < \kappa_i\} = \tau, \]
for some user-specified percentile, $\tau \in (0,100)$, and where $|\cdot|$ is the cardinality.
In other words, the threshold for each sensor, $\kappa_i$, is set as the $\tau$-th percentile of the normalised error scores across the neighbourhood of $i$, on the validation data set.
Unless otherwise stated, we set $\tau = 99$.
In this way, the sensor threshold is based only on its direct neighbourhood, as opposed to the original method which uses the global maximum, and is thus in tune with the local behaviour of the system, and more robust as a percentile.
We refer to the GDN model using this variant of the threshold-based anomaly detection as GDN+.

\subsection*{New Class of Benchmarking Data} \label{sec:simulated}

\begin{figure}
\centering
\includegraphics[width=0.48\textwidth]{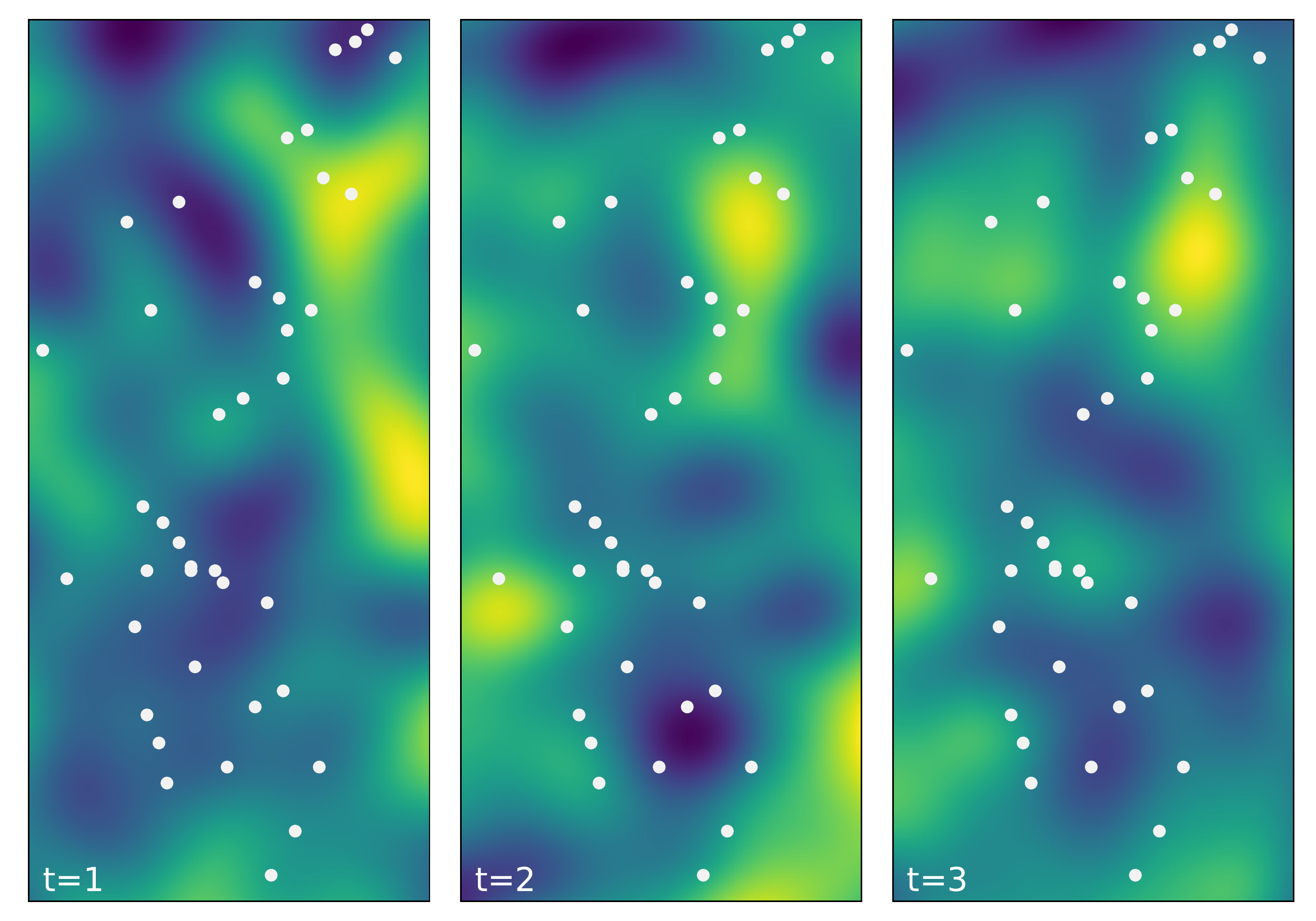}
\caption{\label{fig:xtime} Smooth random Gaussian field used to generate covariate values, $\v X$, in the simulation studies. Examples of the field are shown for $t = 1,2,3$. Sensor locations are shown as white dots.}
\end{figure}

The following is a method for simulating synthetic datasets with persistent anomalies inspired by the statistical models recently used to model river network data \cite{santos2022bayesian}.
Let $\cal S$ denote an arbitrary set of individual spatial locations, with locations $\v s_1, \v s_2,\ldots, \v s_{n} \in \cal S$ chosen by experimental design \cite{buchhorn2022bayesian} or otherwise. Consider a linear mixed model with $n \times 1$ response $\v Y$, and $n \times m$ design matrix $\v X$ of explanatory variables spatially indexed at locations $\v s_1, \v s_2,\ldots, \v s_{n}$, 
\begin{equation}
    \v Y_t = \v \beta_0 + \v X_t \boldsymbol{\beta} + \v Z + \boldsymbol{\epsilon_0}, \quad t=1,\ldots, T, 
\label{eq:linear}
\end{equation}

with time-homogeneous spatially-correlated random effects $\v Z \sim \mathcal{N}(\v 0, \boldsymbol{\Sigma}_{\v Z})$ and vector of independent noise terms $\boldsymbol{\epsilon_0} {\sim} \mathcal{N}(\v 0, \sigma_0^2 \m I)$, yielding $\mbox{Cov\,}[\v Y_t| \v X_t = \v x_t] = \boldsymbol{\Sigma}_{\v Z} + \sigma_0^2 \m I$, where $\m I$ denotes the $n \times n$ identity matrix and $\v \epsilon_0$ is an error term. The covariates $\v X_t$ for $t=1,\ldots, T$ are simulated according to an autoregressive process based on an underlying sequence of independent random fields,
\begin{equation}
    \v X_t = \Sigma^p_{i=0} \varphi_i \widetilde{\v X}_{t-i}, 
    \label{eq:covariates}
\end{equation}
where $p$ is the order of the autoregressive process, and
\[\widetilde{\v X}_t \simiid {\cal N}(\v 0, \v \Sigma_{\v X}), \quad t=1,\ldots, T.\]
Note that other distributions may be used. Above, 
\[(\v \Sigma_{\v X})_{ij} = k(\v s_i, \v s_j') \]
for some covariance kernel $k$. For example,
\begin{equation}
k(\mathbf{s},\mathbf{s}'; \sigma, \alpha) = \sigma^2 \exp\left(-\frac{||\mathbf{s} - \mathbf{s}'||^2}{\alpha}\right),
 \label{eq:coveuc}   
\end{equation}
where $||\cdot ||$ denotes the Euclidean norm, $\sigma^2 > 0$ is the covariance-scaling parameter, and $\alpha \in \bb R$ is the range parameter that controls the rate of decay of correlation between points over distance.
Figure \ref{fig:xtime} illustrates an example of a generated Gaussian random field evolving over time.\\

\begin{figure}
\centering
\includegraphics[width=0.22\textwidth]{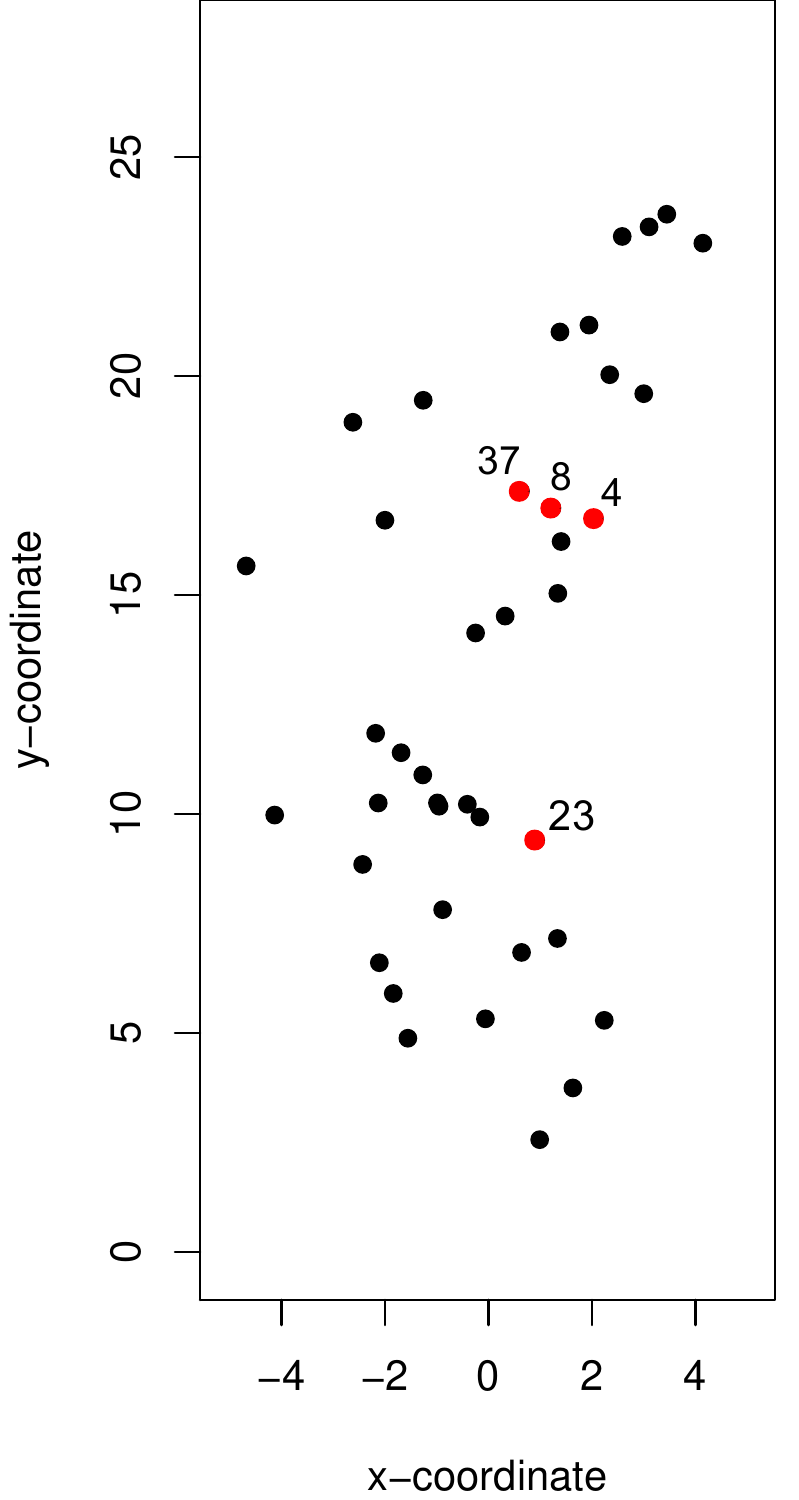}
\includegraphics[width=0.22\textwidth]{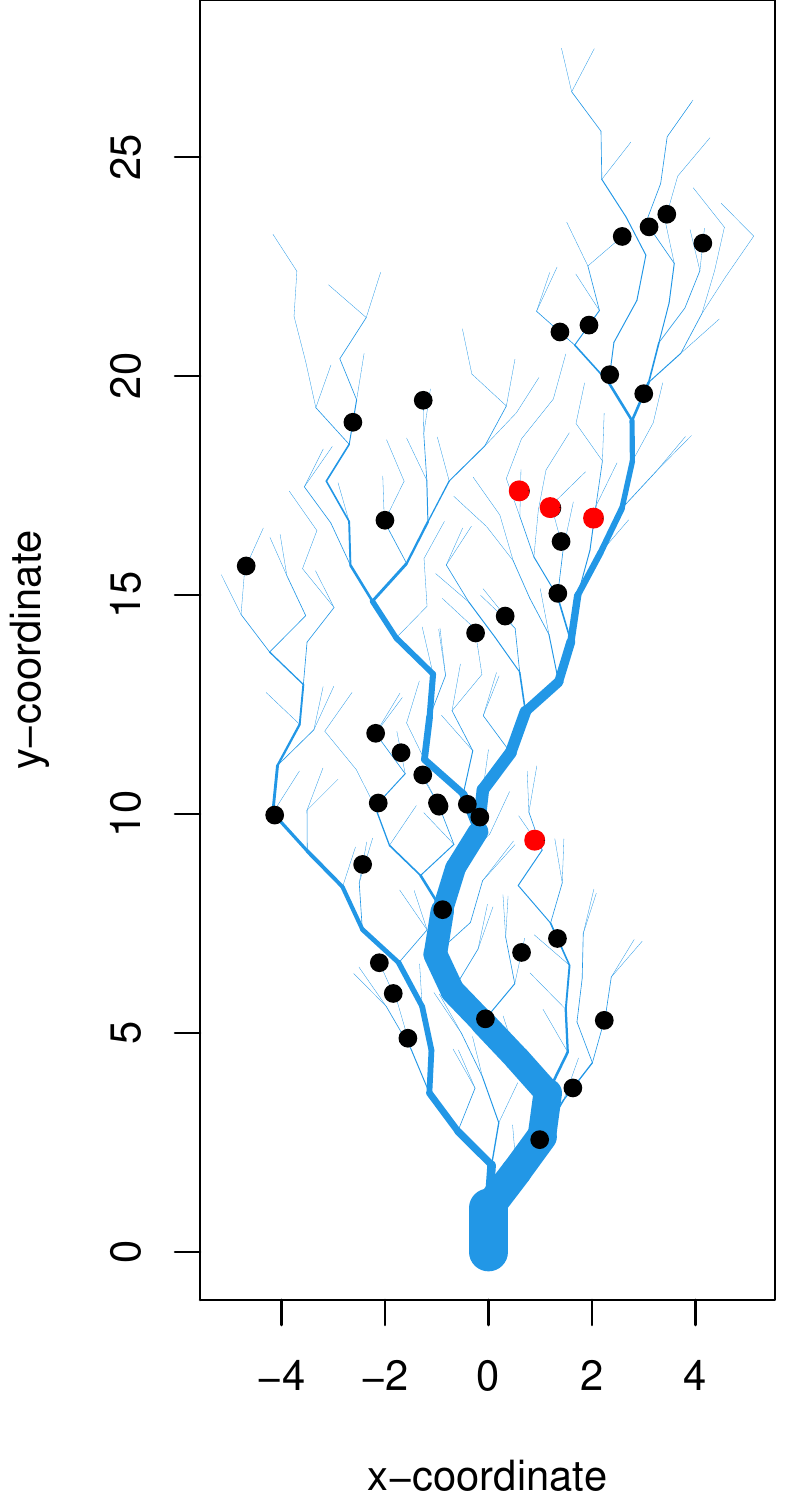}
\caption{\label{fig:locations} {\em SimEuc} site locations across space (left) and {\em SimRiver} site locations along a river network (right). Both simulations use the same (x, y) coordinates for sensor locations. The direction of flow is from top to bottom for SimRiver. Red dots indicate sites for which time series have been shown in Figure \ref{fig:simeuc} and Figure \ref{fig:simriver}.}
\end{figure}

We consider two scenarios in generating simulated data: 1) spatial relationships characterised by Euclidean distance only, where the covariance matrix, $\boldsymbol{\Sigma}_Z$, is constructed via the kernel function given in Equation \ref{eq:coveuc}, and 2) data simulated on a river network, where $\boldsymbol{\Sigma}_Z$ is constructed via the kernel function given in Equation \ref{eq:rivercov}.
Together, this approach provides a variety of simulated data with highly-sophisticated dependency structure, see Figure \ref{fig:locations}.

Two points $\v s_i$, $\v s_j$ on a river network are said to be \emph{flow-connected} if they share water flow, and \emph{flow-unconnected} otherwise. 
We define stream distance, $h_{Riv}(\v s_i,\v s_j)$, as the shortest distance separating $\v s_i$ and $\v s_j$ when travelling {\em along} a given river network.
Tail-up covariance models for river networks, introduced in \cite{ver2010moving}, effectively represent spatial relationships when variables are dominated by flow (e.g. pollutants enter a stream and only impact downstream locations).
By construction, the tail-up covariance function only allows for correlation between flow-connected sites:
\begin{equation}
k_{\rm TU}(\v s_i, \v s_j; \sigma, \alpha) = 
		\omega_{ij} \sigma^2 \exp\left(-\frac{h_{\rm Riv}(\v s_i, \v s_j)}{\alpha}\right) \mathcal{F}_{ij},
\label{eq:rivercov}
\end{equation}
where ${\cal F}_{ij}$ is equal to one if $s_i$ and $s_j$ are \text{flow-connected}, and zero otherwise, and $\omega_{ij}$ is a weighting attributed to each stream segment to account for the upstream branching network structure and ensure stationarity in variance (for full details, see \cite{cressie2006spatial, ver2006spatial}).
The weightings corresponding to each segment may incorporate flow volume, or the area of the catchment, or a proxy such as stream order \cite{shreve1966statistical}.
Note that there are various choices of covariance models. 
Tail-down models allow correlation between both flow-connected and flow-unconnected locations, and may be more suitable for water variables such as temperature, or organisms that can move both upstream and downstream \cite{peterson2003upstream}.
Here we use the exponential function for decay, for further covariance model examples see \cite{ver2010moving}.

Once the base multivariate time series is constructed as above, it is modified to include persistent anomalies as follows. 
Hyperparameters are, $n_{\rm anomaly} \ge 0$, the number of subsequence anomalies and, $\lambda_{\rm anomaly} > 0$, the average length of an anomalous subsequence, for each anomaly type.
In this example, we consider two types of anomalies, high-variability and drift, see Algorithm \ref{algo:generation}.

\begin{algorithm}[!ht]
\SetKwInOut{KwInput}{input}
\SetKwInOut{KwOutput}{output}
\DontPrintSemicolon
\caption{Two-type Anomaly Generation}\label{algo:generation}
\KwInput{Time series data $\m Y = \left[\v y^{(1)}, \ldots, \v y^{(T)}\right]$, number of locations $n$, expected length of each anomaly type $\lambda_{\rm drift}$ and $\lambda_{\rm var}$, number of anomalies, $n_{\rm drift}$ and $n_{\rm var}$, variability anomaly scale $\zeta$, and drift anomaly parameter $\delta$.}

\For{${\rm anomaly} \in \{{\rm drift}, {\rm var}\}$}{
\For{$i=1,\ldots, n_{\rm anomaly}$}{
    Draw location $S \sim  {\rm Uniform}(\{1, 2,\ldots, n\})$\\
    Draw time $t \sim {\rm Uniform}(\{1,\ldots, T \})$\\
    Draw length $L \sim {\rm Poisson}(\lambda_{\rm anomaly})$\\
\uIf{${\rm anomaly} = {\rm drift}$}{
$\v v \leftarrow (\delta, 2\delta, \ldots, L\delta)$ \tcp*{drift}
}\uElse{
$\v v \sim \mathcal{N}(\v 0, \zeta^2 \m I_{L \times L})$ \tcp*{variability}
}
$\v y_{S, t:(t + L)} \leftarrow \v y_{S, t:(t + L)} + \v v$
}
}
\end{algorithm}


\subsection*{Python Package: Graph-Based Neural Network Anomaly Dectection ({\fontfamily{qcr}\selectfont gnnad})}

The Python package {\fontfamily{qcr}\selectfont gnnad} introduced in this paper extends and generalises the research code originally implemented by \cite{deng2021graph}, which is incompatible with newer package dependencies and offers only a command line interface.
The code is refactored to be modular and user-friendly, with a scikit-inspired interface, and extended to include visualisation, data and anomaly generation modules, as well as the GDN+ model extension.
A continuous integration/continuous deployment (CI/CP) pipeline is established with unit testing to ensure that changes to the code are tested and deployed efficiently. 
Comprehensive documentation now accompanying the codebase enhances readability for future developers, facilitating maintenance, reuse, and modification.
Furthermore, rigorous error handling is implemented to improve the software experience.
The software developments have resulted in a more robust, user-friendly and easily distributable package that is available via \url{https://github.com/KatieBuc/gnnad} and the pip repository, {\fontfamily{qcr}\selectfont gnnad}.
See below for example code that shows the process of fitting the GDN+ model.

\begin{lstlisting}[language=Python, basicstyle=\footnotesize]
from sklearn.model_selection import 
... train_test_split
from gnnad.graphanomaly import GNNAD
from gnnad.generate import GenerateGaussian,
... GenerateAnomaly

# generate sample data
gengauss = GenerateGaussian(T=4000, seed=435,
... n_obs=20)
X = gengauss.generate()

# split train test
X_train, X_test = train_test_split(X)

# generate anomalies on test set
anoms = GenerateAnomaly(X_test)
X_test = anoms.generate(anoms.variability, 
... lam = 3, prop_anom = 0.07, seed=45)
X_test = anoms.generate(anoms.drift,
... lam = 11, prop_anom = 0.07, seed=234)
y_test = anoms.get_labels()

# instantiate and fit GDN model object 
model = GNNAD(threshold_type="max_validation",
... topk=6, slide_win=200)
fitted_model = model.fit(X_train, X_test,
... y_test)

# sensor based threshold based on GDN+
pred_label = fitted_model.sensor_threshold_
... preds(tau = 99)

# print evaulation metrics
fitted_model.print_eval_metrics(pred_label)
\end{lstlisting}

\section*{Results}

This section presents a summary of the main findings for anomaly detection using GDN/GDN+ on both simulated and real-world data.
To ensure quality of data, the aim for practitioners is to maximise the ability to identify anomalies of different types, while minimising false detection rates.
We define the following metrics in terms of true positive (TP), true negative (TN), false positive (FP) and false negative (FN) classifications. 
In other words, the main priority is to minimise FN, while maintaining a reasonable number of FP, such that it is not an operational burden to check the total number of positive flags.
Accordingly, we use recall, defined by $\frac{{\rm TP}}{{\rm TP} + {\rm FN}}$, to evaluate the performance on the test set and to select hyperparameters.
That is, the proportion of actual positive cases that were correctly identified by the model(s).
We also report the performance using precision $\big(\frac{{\rm TP}}{{\rm TP} + {\rm FP}}\big)$, accuracy $\big(\frac{{\rm TP} + {\rm TN}}{{\rm TP} + {\rm TN} + {\rm FP} + {\rm FN}}\big)$ and specificity $\big(\frac{{\rm TN}}{{\rm TN} + {\rm FP}}\big)$.

To evaluate model performance, three existing anomaly detection models are used as benchmarks: 1. The naive (random walk) Autoregressive Integrated Moving Average Model (ARIMA) prediction model from \cite{hyndman2018forecasting, leigh2019framework}; 2. HDoutliers \cite{wilkinson2017visualizing}, an unsupervised algorithm designed to identify anomalies in high-dimensional data, based on a distributional model that allows for probability assignment to an anomaly; and 3. DeepAnT \cite{munir2018deepant}, an unsupervised, deep learning-based approach to detecting anomalies in time series data.

\begin{table}[h!]
\hrule \vspace{0.1cm}
\caption{\label{tab:datacount}Details of the three data sets used in the case studies.}
\centering
\begin{tabledata}{lrrr} 
\header Dataset & \#Train & \#Test & \%Anomalies\\ 
\row {\em SimEuc} & $3,000$ & $1,000$ & $13.2$ \\ 
\row {\em SimRiver} & $3,000$ & $1,000$ & $13.2$\\
\row Herbert & $12,745$ & $3,499$ & {\bf 58.0}\\ 
\end{tabledata}
\end{table}

\subsection*{Simulation Study: Benchmark Data}

Data are generated using the linear-mixed model described in Equation \ref{eq:linear}, with differing spatial dynamics: {\em SimEuc} where the random effect, $\v Z$, is characterised by Euclidean distance only, and {\em SimRiver} where $\v Z$ simulates complex river network dynamics \citep{ver2014ssn}, using the same site locations and covariate values, $\v X$.
Detecting anomalies that involve multiple consecutive observations is a difficult task that often requires user intervention, and is the focus of this study.
We consider scenarios with drift and high-variability anomaly types, which together contaminate $13.2\%$ of the test data, given $n_{\rm drift}=5, \lambda_{\rm drift}=11, n_{\rm var}=24, \lambda_{\rm var}=3$, see Table~\ref{tab:datacount}.

\begin{figure}
\centering
\includegraphics[width=0.48\textwidth]{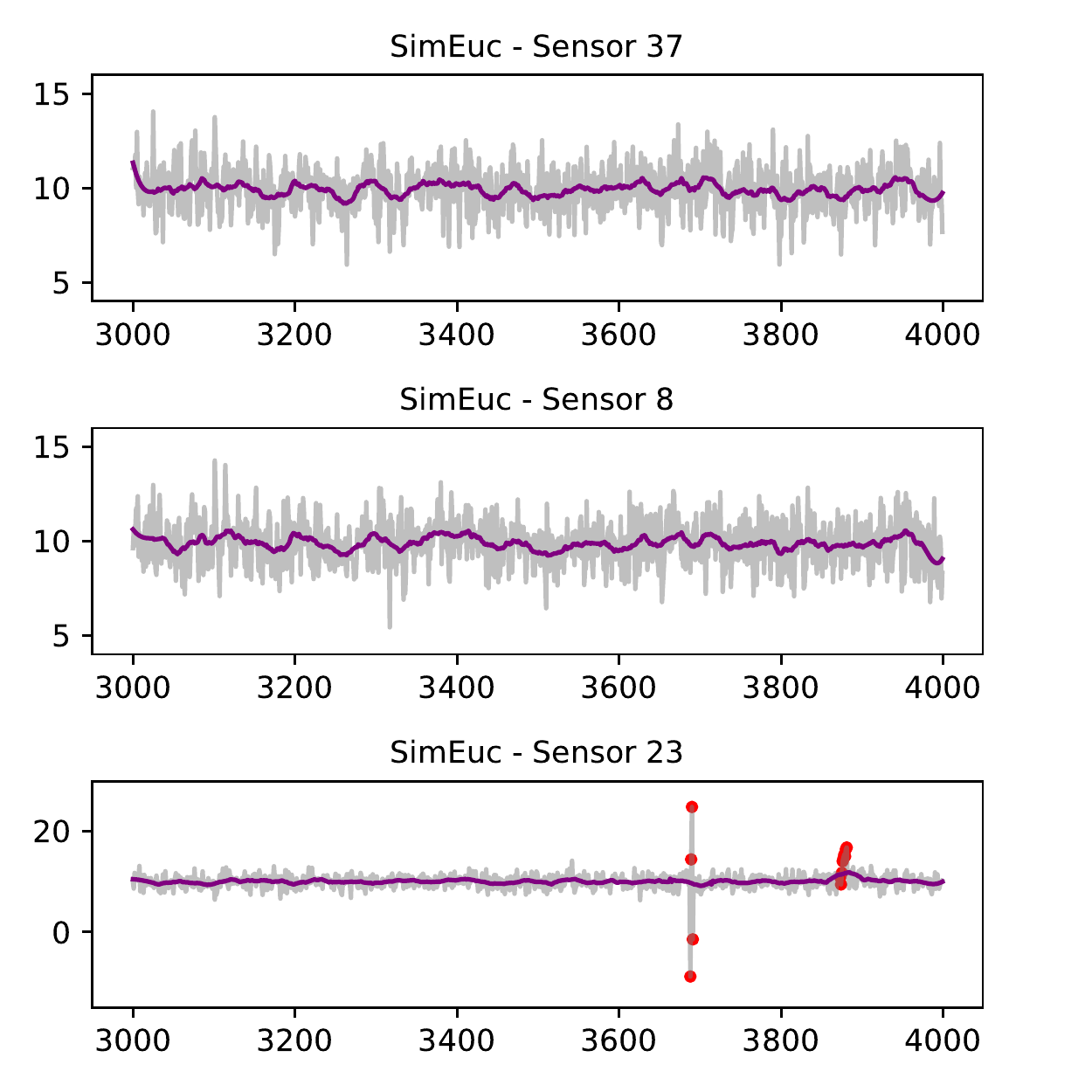}
\caption{\label{fig:simeuc} A selection of time series from the {\em SimEuc} data set; sensor 8 and sensor 37 separated by a short Euclidean distance share high correlation (Pearson`s coefficient of 0.65). A Savitzky-Golay filter smoothens the time series (purple line). On the bottom, sensor 23 illustrates high-variability and drift anomalies, consecutively (red dots).}
\end{figure}

\begin{figure}
\centering
\includegraphics[width=0.48\textwidth]{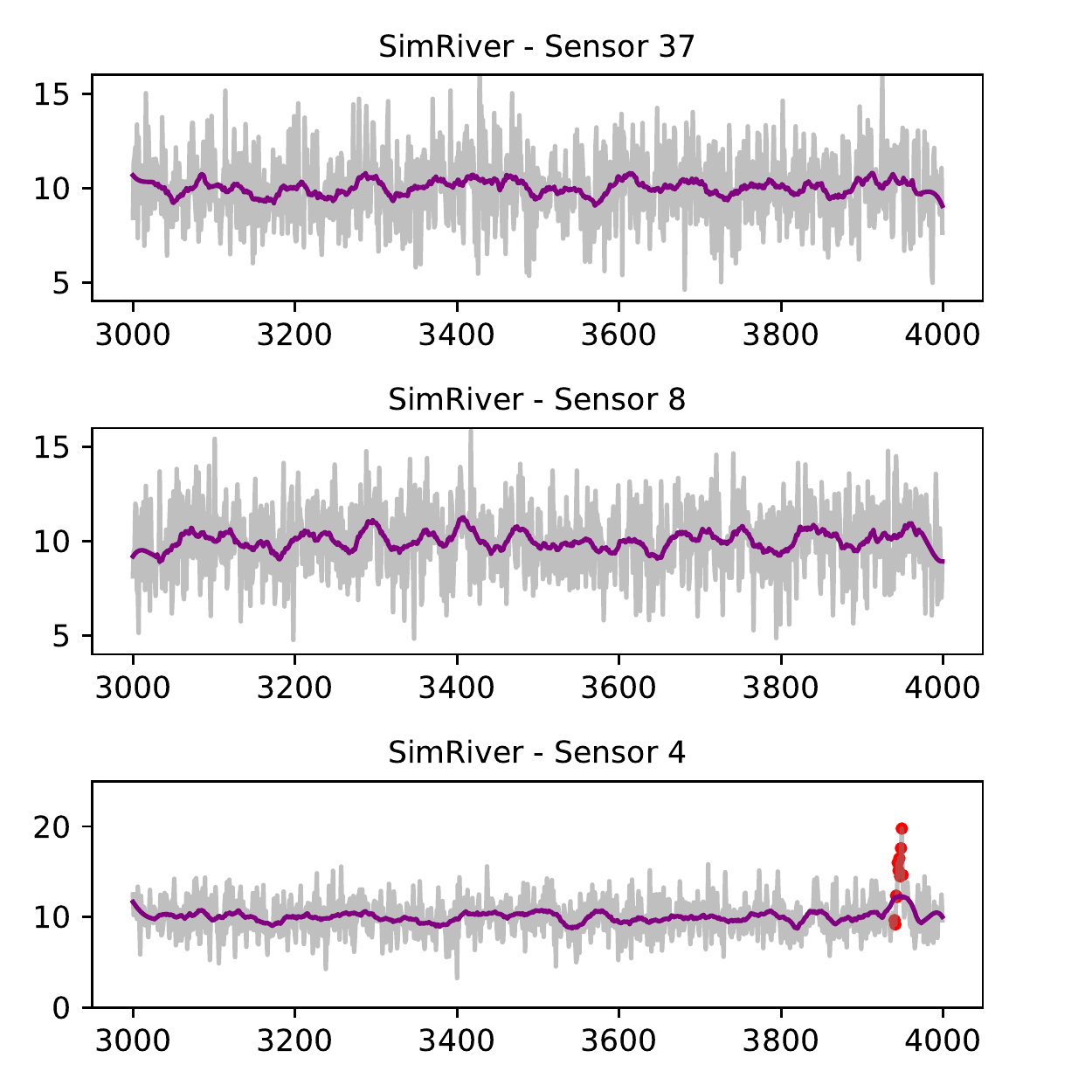}
\caption{\label{fig:simriver}  A selection of series from the {\em SimRiver} data set; sensor 8 and sensor 37 are not flow-connected sites, and share low correlation (Pearson`s coefficient of 0.07). Drift anomalies (red dots) are shown in data from sensor 4. A Savitzky-Golay filter is used to smoothen the time series for visual representation (purple line).}
\end{figure}

Figure \ref{fig:simeuc} visualises aspects of the {\em SimEuc} dataset, where sensor 37 and sensor 8 are in close (Euclidean) proximity, resulting in a high correlation between the locations (0.65), as anticipated.
Note that sensor 23 exhibits anomalous behavior, high-variability and drift, consecutively, over time.
Compared to the {\em SimRiver} dataset, shown in Figure \ref{fig:simriver}, we note how the time series from sensor 37 and sensor 8 are no longer strongly correlated (0.07), despite their close proximity, as they are not flow-connected in the simulated river network.

\begin{table}[h!]
\hrule \vspace{0.1cm}
\caption{\label{tab:performance}Anomaly detection performance in terms of recall (\%), precision (\%), accuracy (\%), and specificity (\%) of GDN and its variants and baseline methods for the simulation study.}
\centering
\begin{tabledata}{llrrrr}
\header Data & Model & Rec & Prec & Acc & Spec \\
\row SimEuc & HDoutliers & {\color{red}{\bf 0.0}} & {\color{red}{\bf 0.0}} & 86.8 & 100.0 \\
\row & ARIMA & {\bf 88.6} & 14.4 & 28.6 & 19.4 \\
\row & DeepAnT & 3.8 & 11.9 & 83.5 & 95.7 \\
\row & GDN & 83.3 & 55.3 & 88.9 & 89.7 \\
\row & GDN+ & \underline{85.6} & 48.1 & 85.9 & 85.9 \\
\hline
\row SimRiv & HDoutliers & {\color{red}{\bf 0.0}} & {\color{red}{\bf 0.0}} & 86.8 & 100.0 \\
\row & ARIMA & {\bf 91.7} & 14.2 & {\color{red}{\bf 25.3}} & {\color{red}{\bf 15.1}} \\
\row & DeepAnT & {\color{red}{\bf 0.8}} & {\color{red}{\bf 9.1}} & 85.9 & 98.8 \\
\row & GDN & 72.7 & 54.2 & 88.3 & 90.6 \\
\row & GDN+ & \underline{78.0} & 43.1 & 83.5 & 84.3 \\
\hline
\end{tabledata}
\end{table}

Table \ref{tab:performance} shows the performance of the different anomaly detection methods. 
The best performance in terms of recall is highlighted in bold, while the second-best performance is underlined.
We observe that GDN/GDN+ outperforms most other models in terms of recall, which is the fraction of true positives among all actual positive instances.
Specifically, GDN has a recall score of 83.3\% on {\em SimEuc} and 72.7\% on SimRiv, while GDN+ has the second highest recall score of 85.6\% on {\em SimEuc} and 78.0\% on SimRiv.
Although ARIMA performed best in terms of recall, the high percentage of detected anomalies, 81.6\% and 85.6\%, is impractical to use (discussed below) and results in low accuracy scores of 28.6\% and 25.3\%, respectively.
HDoutliers did not flag any time point as anomalous in any test case.
DeepAnT tends to classify most samples as negative, resulting in a low recall score.
These results suggest that GDN/GDN+ are best able to detect a high proportion of actual anomalies in the datasets.

\begin{figure}
\centering
\includegraphics[width=0.45\textwidth]{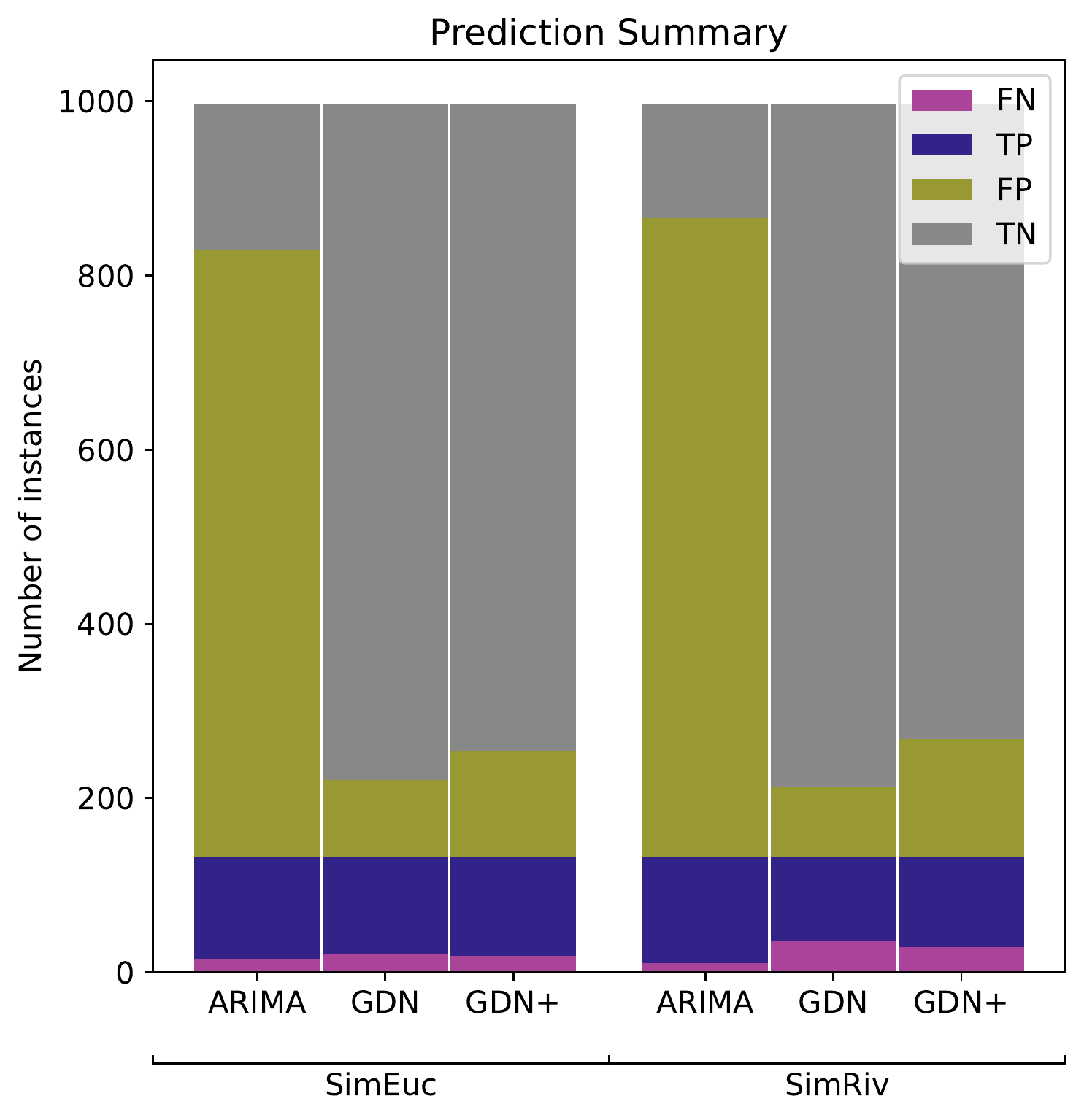}
\caption{\label{fig:confusion} Anomaly detection performance of ARIMA, GDN, and GDN+ for the simulation study, in terms of true positive (TP), true negative (TN), false positive (FP) and false negative (FN).}
\end{figure}

Figure \ref{fig:confusion} shows the classifications of anomalies in the simulation study.
For reasons mentioned, recall is the performance metric of interest, but we also consider the trade-off between recall and precision.
Lower precision means that the model may also identify some normal instances as anomalies, leading to false positives. 
In the context of river network anomaly detection, FP may be manually filtered, but it is critical to minimise FN.
Note that GDN+ outperforms GDN in minimising the FN count, but at the cost of increasing FP, in both data sets.
Such a trade-off is acceptable and considered an improvement in this context.
Conversely, while ARIMA has the highest recall score, the number of FP classifications is impractical for practitioners to deal with (>70\% of the test data).
We also note that drift anomalies are harder to detect than high-variability anomalies, with drift as the majority of FN counts, in all cases.

The authors of the GDN model demonstrated its efficacy in detecting anywhere-within-system failures at time $t$ by applying a threshold to all sensors within a system.
However, the use of sensor-based thresholds in GDN+ has the advantage of indicating anomalies at the individual sensor level.
In the context of monitoring river networks, it is crucial to identify the anomalous sensor, $i$, at a given time $t$.
The percentage of true positives detected at the correct sensor, $i$, using the sensor-based anomaly threshold, $A_i(t)$, in GDN+, was 92\% and 89\% for {\em SimEuc} and SimRiver, respectively.
Similarly, the rate of true positives detected in the neighbourhood of the correct sensor $i$ were 96\% and 91\%, respectively.
This granularity of information is essential for large networks consisting of independent sensors that are separated by significant spatial distances, where the cost of time and travel for sensor replacement or maintenance is substantial.

\subsubsection*{Replication Study}

This section explores the anomaly detection performance of GDN/GDN+ across multiple simulated data sets. The approach is as follows. First, ten new sets of spatial sampling locations are created, and for each set a Gaussian random field evolving over time is simulated, as per Equation \ref{eq:covariates}.
For each set of locations, we again consider both the Euclidean spatial characterisation (\textit{SimEuc}), and the river network spatial characterisation (\textit{SimRiver}), yielding a total of 20 benchmark data sets.
In the first case, we use the Euclidean covariance model in Equation \ref{eq:coveuc}, parameterised by, $\sigma^2 \in [1,5]$, and, $\alpha \in [5,15]$, with independent noise parameter, $\sigma_0^2 \in [0, 1]$, and regression parameters $\beta_0 \in [1,10]$, and, $\beta_1 \in [1,10]$, for the linear model in Equation \ref{eq:linear}.
The values of the parameters are chosen uniformly at random. 
The Tail-up covariance model in Equation \ref{eq:rivercov} is used in the second case, parameterised as above.

Then, anomalies are generated with the following parameters: drift $\delta \in [3,6]$, variability $\zeta \in [12,15]$, length of anomalies $\lambda_{\rm drift} \in [5,10]$, $\lambda_{\rm var} \in [2,10]$, and the number of anomalies, $n_{\rm drift}\in [50,100]$, $n_{\rm var}\in [50,100]$ (see Algorithm \ref{algo:generation}).
Across all simulations, the size of the data set is fixed to have length, $T=4000$, and number of sensors, $n=40$.

\begin{figure}
\centering
\includegraphics[width=0.48\textwidth]{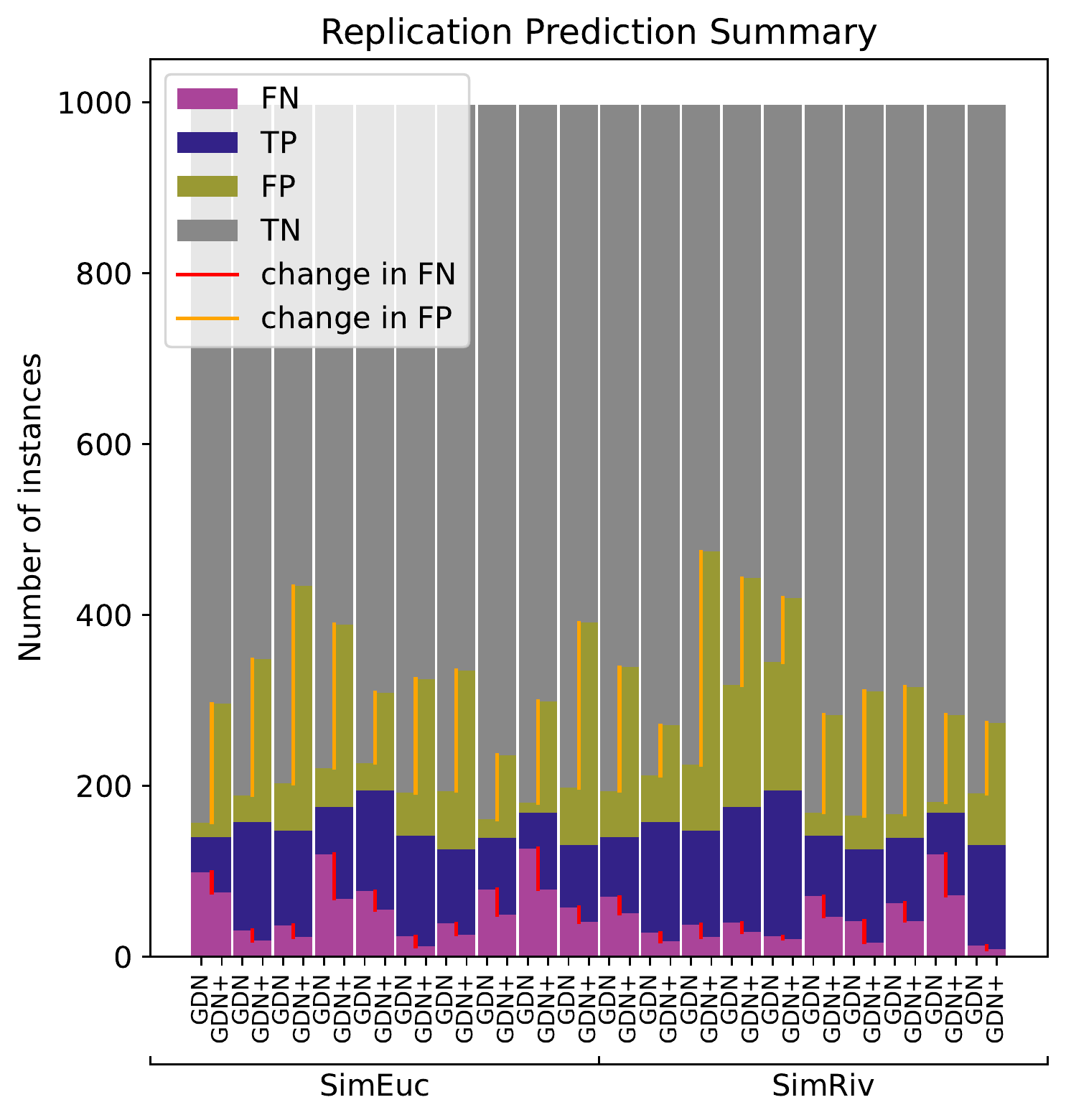}
\caption{\label{fig:replication} Anomaly detection performance of GDN, and GDN+ across the twenty simulated benchmarks in the replication study. Note that GDN+ consistently decreases false negatives (red line) in every case, but also increases false positives (orange line).}
\end{figure}

Figure \ref{fig:replication} illustrates the anomaly detection performance for GDN and GDN+, run on each data set with sliding window length $w = 3$, and Top-K hyperparameter $K=5$.
Note that the total number of anomalies can be seen at the bar height of TP+FN.
In every scenario, GDN+ improves the FN count (red line), but at the cost of an increased TP count (orange line). Whether such a tradeoff is tolerable depends on how critical it is in practical scenarios that true anomalies are successfully detected. Note that performance varies from one scenario to the next. Nevertheless, despite the simulated datasets being extremely noisy and  complex, GDN and GDN+ appear to succeed in successful anomaly detection when other methods cannot.

\subsection*{Case Study: Herbert River}

\begin{figure}
\centering
\includegraphics[width=0.48\textwidth]{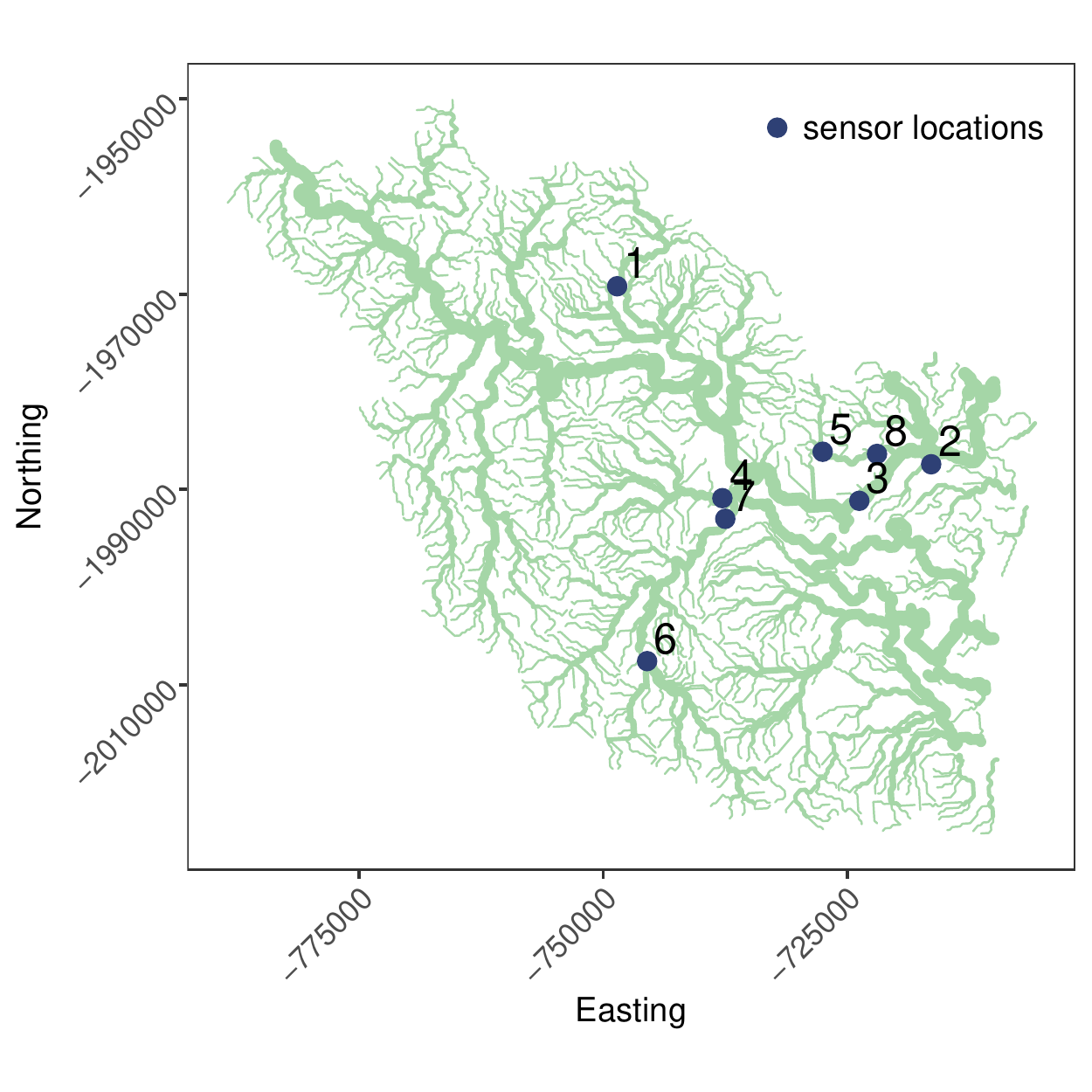}
\caption{\label{fig:herbert_network} The Herbert river system and sensor locations.}
\end{figure}

\begin{figure}
\centering
\includegraphics[width=0.48\textwidth]{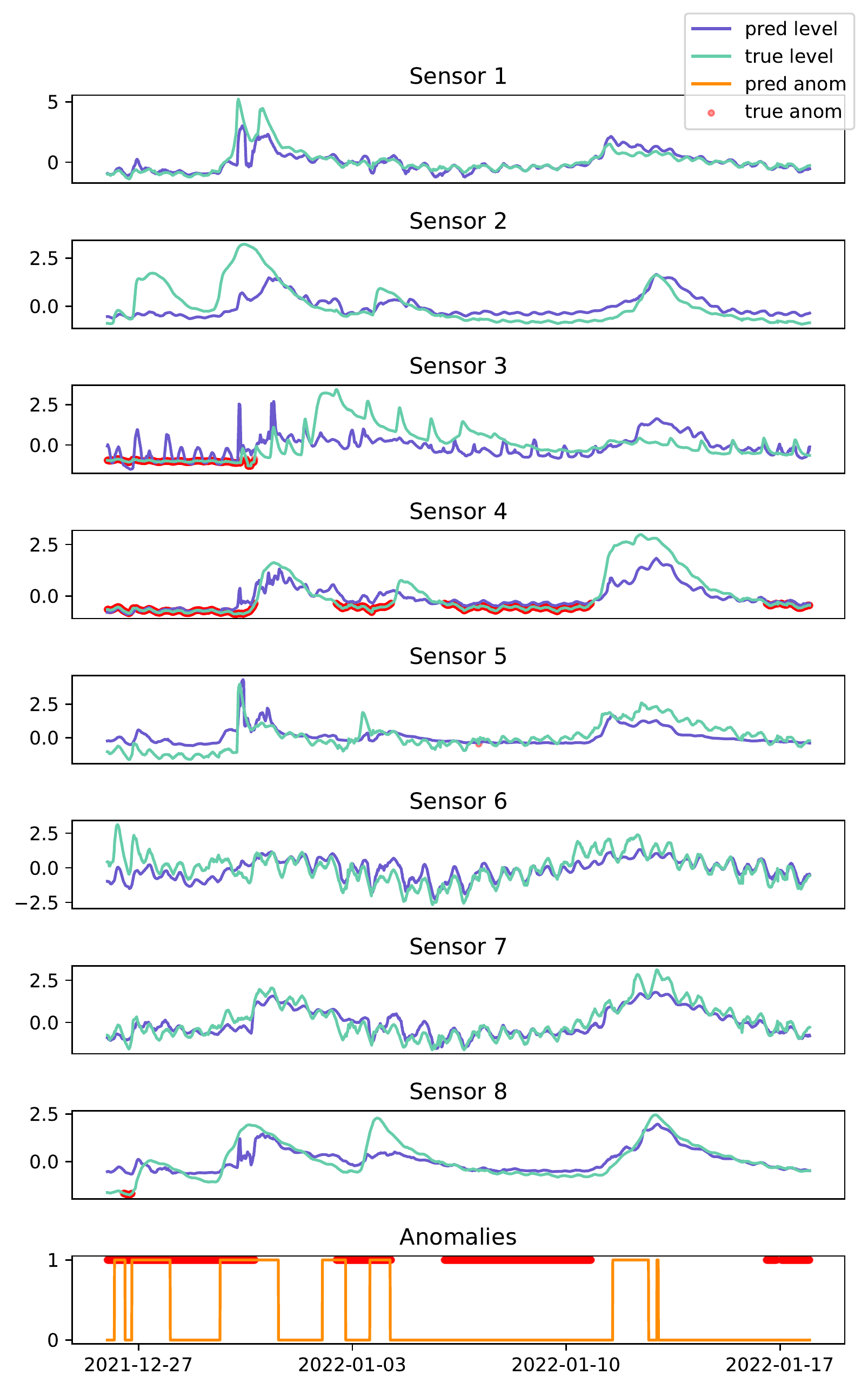}
\caption{\label{fig:herbert} Test data of water level collected from sensors across the Herbert river (light blue), and the corresponding predictions from the GDN model (dark blue). Actual anomalies (red dots) are shown, along with the predicted anomalies (orange line) on the bottom.}
\end{figure}

This case study examines water-level data collected from eight sites located across the Herbert river, a major river system located in the tropical region of Australia, as shown in Figure \ref{fig:herbert_network}.
The time series data is highly non-stationary, characterised by river {\em events} caused by abnormal rainfall patterns, with some coastal sites exhibiting shorter periodicity trends which can be attributed to tidal patterns, see Figure \ref{fig:herbert}.
The spatial relationships are complex, and depend on the surrounding water catchment areas, spatial rainfall patterns, dams, and other impediments.
In-situ sensors are prone to various anomalies such as battery failure, biofouling (accumulation of microorganisms, plants, algae, or small animals), and damage.
In some cases, anomalies can manifest as the absence of a water event (i.e., flatlining) rather than the presence of abnormal time series patterns (i.e., spikes, variability, drift).
In real-world scenarios, anomalies can persist for extended periods, and resolving them may require traveling to remote locations to inspect and repair sensors.
As seen in Figure \ref{fig:herbert}, anomalies at time $t$ are largely attributed to Sensor 4, which was out of water for long periods of time.

The Herbert river is a challenging data set for all of the anomaly detection models, due to the sparse placement of sensors across the network, i.e., fewer sensors at greater distances apart resulting in weaker spatial relationships, and the test set contains a high proportion of anomalies (58\%; see Table \ref{tab:datacount}).
GDN applied to the real-world dataset yields a recall of $29.2\%$, with GDN+ improving recall to $34.8\%$, see Table \ref{tab:herbertperformance}.
Model performance suffers primarily due to the failure to detect the large anomaly spanning across 2022-01-10 in Figure \ref{fig:herbert}.
This may be attributed to the learned graph relationships being characterised by river events in the training data, and without such events, it is difficult to identify when a sensor is flat-lining.
However, the model successfully identified anomalies spanning across 2021-12-27 and 2022-01-03, which coincided with river events. 

\begin{table}[h!]
\hrule \vspace{0.1cm}
\caption{\label{tab:herbertperformance}Anomaly detection performance in terms of recall (\%), precision (\%), accuracy (\%), and specificity (\%) of GDN and its variants and baseline methods for the Herbert river case study.}
\centering
\begin{tabledata}{llrrrr}
\header Data & Model & Rec & Prec & Acc & Spec \\
\row Herbert & HDoutliers &{\color{red}{\bf 0.0}} & {\color{red}{\bf 0.0}}& 42.0 & 100.0 \\
\row & ARIMA & 30.5 & 62.9 & 51.3 & 77.5 \\
\row & DeepAnT & {\color{red}{\bf 1.6}} & 39.7 & 44.0 & 97.0 \\
\row & GDN & 29.2 & 60.6 & 50.1 & 76.2 \\
\row & GDN+ & \underline{34.8} & 59.2 & 50.4 & 70.0 \\
\row & GDN++ & {\bf 100.0} & 80.7 & 86.7 & 70.0 \\
\hline
\end{tabledata}
\end{table}

\begin{figure}
\centering
\includegraphics[width=0.4\textwidth]{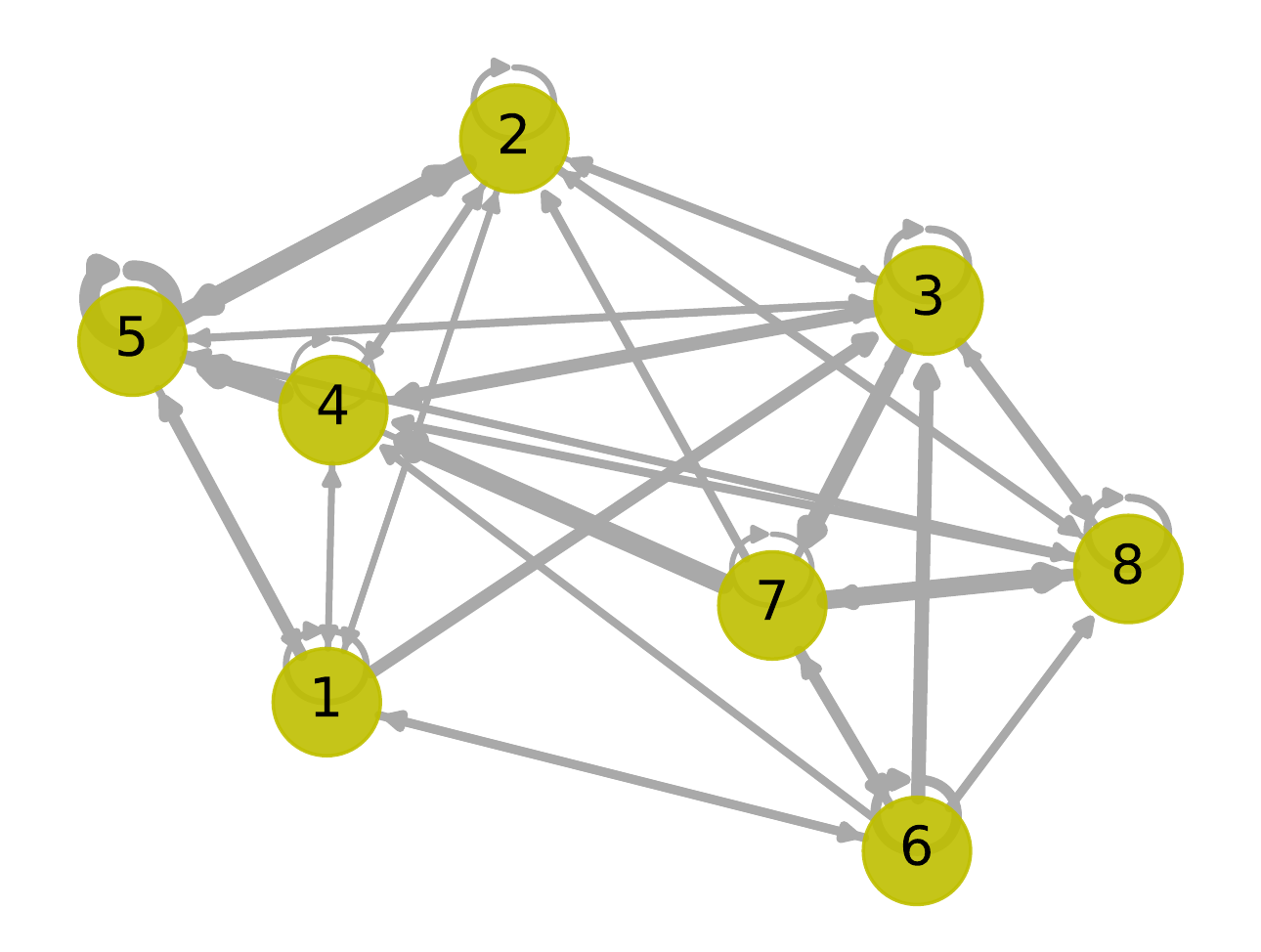}
\caption{\label{fig:attention} Graph with learned adjacency matrix, $A$. The edge weightings are determined by $\alpha_{ij}$, which indicates attention and depend on model input, ${\rm X}^{(t)}$.}
\end{figure}

Figure \ref{fig:attention} shows the learned graph adjacency matrix, $A$. 
Sensor 1, separated geographically by a large distance from the other sensors, has weaker relationships with the other nodes.
Interestingly, the attention weights indicate that sensor 3 is strongly influenced by sensor 6, with tidal patterns from sensor 6 being evident in the predictions of sensor 3.
Large error scores indicating an anomaly spanning across 2021-12-27 are primarily observed in sensor 2, impacted by anomalous sensors 3 and 4, where the predicted values are low.
However, due to the small network of sensors in this case study, it is difficult to determine which sensor had the anomaly.

Adjusting the threshold in the GDN+ model can only enhance performance to a limited extent, since some anomalies may have insignificant error scores due to the underlying time series model.
For instance, the anomaly spanning across 2022-01-10 has negligible error scores because the prediction model was performing well and no river events had occurred, making it challenging to detect the flat-lining type of anomaly in this particular scenario.
Therefore, relying solely on the model may not be sufficient in practice, and we recommend implementing some basic expert rules.
We introduce another model variant GDN++, by applying a simple filter to the time series ensuring that all values are positive, used in conjunction with adjusting the threshold calculation (GDN+). 
That is, an anomaly at time, $t$, on sensor, $i$, is flagged if, $\max \{\bb I \{y_i^{(t)} < 0\}, A_i(t)\} = 1.$
GDN++ successfully detects all anomalies.

\section*{Discussion}
Multivariate anomaly detection and prediction models for spatio-temporal sensor data have the potential to transform water quality observation, modelling, and management \cite{ver2010moving, leigh2019framework}.
The provision of trustworthy sensor data has four major benefits: 1. It enables the production of finer-scale, reliable and more accurate estimates of sediment and nutrient loads, 2. It provides real-time feedback to landholders and managers, 3. It guides compliance with water quality guidelines, and 4. It allows for the assessment of ecosystem health and the prioritisation of management actions for sustaining aquatic ecosystems.
However, technical anomalies in the data provided by the sensors can occur due to factors such as low battery power, biofouling of the probes, and sensor miscalibration.
As noted by \cite{leigh2019framework}, there are a wide variety of anomaly types present within in-situ sensor data (e.g., high-variability, drift, spikes, shifts).
Most anomaly detection methods used for water quality applications tend to target specific anomalies, such as sudden spikes or shifts \cite{talagala2019feature, hill2010anomaly, ba2015water}.
Detecting persistent anomalies, such as sensor drift and periods of abnormally high variability, remains very challenging for statistical and machine learning research.
Such anomalies are often overlooked by distance and kernel based methods, yet must be detected before the data can be used, because they confound the assessment of status and trends in water quality.
Understanding the relationships among, and typical behaviours of, water quality variables and how these differ among climate zones is thus an essential step in distinguishing anomalies from real water quality events.

We investigated the graph-based neural network model, GDN \cite{deng2021graph}, for its ability to capture complex interdependencies between different variables in a semi-supervised manner.
As such, GDN offered the ability to capture deviations from expected behaviour within a high-dimensional setting.
We developed novel bench-marking data sets for subsequence anomaly detection (of variable length and type), with a range of spatio-temporal complexities, inspired by the statistical models recently used for river network data \cite{santos2022bayesian}.
Results showed that GDN tended to outperform the benchmarks in anomaly detection.
We developed a model extension, GDN+, by adjusting the threshold calculation.
GDN+ was shown to further improve performance.
A replication study with multiple benchmarking data sets demonstrated consistency in these results.
Sensor-based thresholds also proved useful in terms of identifying which neighbourhood the anomaly originated from in the simulation study.

We used a real-world case study of water level in the Herbert river, with non-stationary time series characterised by multiple river events caused by abnormal rainfall patterns.
In this case, most of the anomalies appeared as flat-lining, due to the river drying out.
Considering an individual time series, such anomalies may not appear obvious, as it is the failure to detect river events (in a multivariate setting) that is indicative of an anomalous sensor.
Despite the challenges in the data, GDN+ was shown to successfully detect technical anomalies when river events occurred, and combined with a simple expert-based rule, GDN++, all anomalies were successfully detected.

There are two recent methodological extensions to the GDN approach.
First, the Fused Sparse Autoencoder and Graph Net \citep{han2022learning} which extends the GDN approach by augmenting the prediction-based loss with a reconstruction-based term arising from the output of a sparse autoencoder, and a further extension that allows the individual sensors to have multivariate data.
Second, a probabilistic (latent variable) extension is trained using variational inference \citep{chen2022deep}.
Since these were published contemporaneously to the present research, and only the latter provided accompanying research code, these approaches were not considered in the paper.
Future extensions of this work could consider incorporating the above methodological extensions, as well as developing on the existing software package.

Other applications could consider the separation of river events from technical anomalies.
In terms of interpretability, since the prediction model aggregates feature data from neighbouring sensors, anomalous data can affect the prediction of any other sensor within the neighbourhood.
Therefore, large error scores originating from one sensor anomaly can infiltrate through to the entire neighbourhood, and impairs the ability to attribute an anomaly to a sensor.
The extensive body of literature on signal processing for source identification has the potential to inspire future solutions in addressing this issue \citep{scharf1991statistical, telford1990applied}.

In summary, this work extends and examines the practicality of the GDN approach when applied to an environmental monitoring application of major international concern, with complex spatial and temporal interdependencies.
Successfully addressing the challenge of anomaly detection in such settings can facilitate the wider adoption of in-situ sensors and could revolutionise the monitoring and management of air, soil, and water.

\section*{Conclusions}
This work studied the application of Graph Deviation Network (GDN) based approaches for anomaly detection on the challenging setting on river network data, which often feature sensors that generate high-dimensional data with complex spatio-temporal relationships.
We introduced alternative defection criteria for the model (GDN+/GDN++), and their practicality  was explored on both real and simulated benchmark data.
The findings indicated that GDN and its variants were effective in correctly (and conservatively) identifying anomalies.
Benchmark data were generated via an approach that was also introduced in this paper, along with open-source software, and may serve useful in the development and testing of other anomaly-detection methods. 
In short, we found that graph neural network based approaches to anomaly detection offer a flexible framework, able to capture and model non-standard, highly dynamic, complex relationships over space and time, with the ability to flag a variety of anomaly types.
However, the task of anomaly detection on river network sensor data remains a considerable challenge.

\subsection*{Author contributions}
KB; Investigation, Formal Analysis, Methodology, Software, Validation, Visualisation, Writing – Original Draft Preparation. RS; Methodology, Supervision, Writing – Review \& Editing. KM; Funding Acquisition, Conceptualisation, Supervision, Writing – Review \& Editing. ES; Data Curation, Visualisation, Writing – Review \& Editing. \\


\subsection*{Grant information}
This work was supported by the Australian Research Council (ARC) Linkage Project (LP180101151) titled ``Revolutionising water-quality monitoring in the information age". Case study data were provided by the Department of Environment and Science, Queensland, and is available as part of the Python package {\fontfamily{qcr}\selectfont gnnad}.


\subsection*{Acknowledgements}
\L ukasz Mentel; Software. Cameron Roberts; Data Curation. James Mcgree; Writing – Review \& Editing. 


{\small\bibliographystyle{unsrtnat}
\bibliography{sample}}

\bigskip





\end{document}